\documentclass[conference]{IEEEtran}
\IEEEoverridecommandlockouts
\usepackage{cite}
\usepackage{amsmath,amssymb,amsfonts}
\usepackage{algorithm}
\usepackage{algorithmic}
\usepackage{graphicx}
\usepackage{textcomp}
\usepackage{xcolor}
\usepackage{cleveref}
\usepackage{microtype}
\usepackage{bm}
\usepackage{epstopdf} 
\usepackage{hyperref}
\usepackage{graphicx}
\usepackage{subfigure}
\usepackage{booktabs} 
\usepackage{amsmath}
\usepackage[utf8]{inputenc}
\usepackage{tikz}
\usepackage{bm}
\usetikzlibrary{positioning,fit,calc}
\usetikzlibrary{arrows}
\usepackage{verbatim}
\usepackage[numbers]{natbib}

\DeclareMathOperator*{\argmin}{arg\,min}

\newcommand*{\connectorHTextBlue}[4][]{
  \draw[#1] (#3) -| node [above,midway, text=blue] {ASC($\mathcal{L}$)} ($(#3) !#2! (#4)$) |- (#4);
}

\newcommand*{\connectorVTextGreen}[4][]{
  \draw[#1] (#3) |- node [left,midway, text=green] {$\mathcal{L}^2_{t}$} ($(#3) !#2! (#4)$) -| (#4);
}

\newcommand*{\connectorHTextGreen}[4][]{
  \draw[#1] (#3) -|  ($(#3) !#2! (#4)$) |-node [left, midway, text=green] {$\mathcal{L}^1_{t}$} (#4);
}

\newcommand*{\connectorHTextGreenRight}[4][]{
  \draw[#1] (#3) -|  ($(#3) !#2! (#4)$) |-node [right, midway, text=green] {$\mathcal{L}^3_{t}$} (#4);
}

\newcommand*{\connectorVTextRed}[4][]{
  \draw[#1] (#3)  |-  ($(#3) !#2! (#4)$) -|node [above, xshift = 0.5 cm, midway, text=red] {$\mathcal{L}_{t+1}$} (#4);
}

\newcommand*{\connectorVTextRedRight}[4][]{
  \draw[#1] (#3)  |-  ($(#3) !#2! (#4)$) -|node [above, xshift = -0.4 cm, midway, text=red] {$\mathcal{L}_{t+1}$} (#4);
}

\newcommand*{\connectorHTextBluea}[4][]{
  \draw[#1] (#3) -| node [above,midway, text=blue] {ASC($\bm\theta$)} ($(#3) !#2! (#4)$) |- (#4);
}

\newcommand*{\connectorVTextGreena}[4][]{
  \draw[#1] (#3) |- node [left,midway, text=green] {$\bm\theta^2_{t}$} ($(#3) !#2! (#4)$) -| (#4);
}

\newcommand*{\connectorHTextGreena}[4][]{
  \draw[#1] (#3) -|  ($(#3) !#2! (#4)$) |-node [left, midway, text=green] {$\bm\theta^1_{t}$} (#4);
}

\newcommand*{\connectorHTextGreenRighta}[4][]{
  \draw[#1] (#3) -|  ($(#3) !#2! (#4)$) |-node [right, midway, text=green] {$\bm\theta^3_{t}$} (#4);
}

\newcommand*{\connectorVTextReda}[4][]{
  \draw[#1] (#3)  |-  ($(#3) !#2! (#4)$) -|node [above, xshift = 0.5 cm, midway, text=red] {$\bm\theta_{t+1}$} (#4);
}

\newcommand*{\connectorVTextRedRighta}[4][]{
  \draw[#1] (#3)  |-  ($(#3) !#2! (#4)$) -|node [above, xshift = -0.4 cm, midway, text=red] {$\bm\theta_{t+1}$} (#4);
}

\usepackage{hyperref}

\def\BibTeX{{\rm B\kern-.05em{\sc i\kern-.025em b}\kern-.08em
    T\kern-.1667em\lower.7ex\hbox{E}\kern-.125emX}}
\begin{document}

\title{Private Dataset Generation Using Privacy Preserving Collaborative Learning
}

\author{\IEEEauthorblockN{1\textsuperscript{st} Amit Chaulwar}
\IEEEauthorblockA{\textit{Electrical and Computer Engineering} \\
\textit{Technische Hochschule Ingolstadt}\\
Ingolstadt, Germanyy \\
amittchaulwar@gmail.com}
}

\maketitle

\begin{abstract}
With increasing usage of deep learning algorithms in  many application, new research questions related to privacy and adversarial attacks are emerging. However, the deep learning algorithm improvement needs more and more data to be shared within research community. Methodologies like federated learning, differential privacy, additive secret sharing provides a way to train machine learning models on edge without moving the data from the edge. However, it is very computationally intensive and prone to adversarial attacks. Therefore, this work introduces a privacy preserving FedCollabNN framework for training machine learning models at edge, which is computationally efficient and robust against adversarial attacks. The simulation results using MNIST dataset indicates the effectiveness of the framework.
\end{abstract}


\section{Introduction}
\label{Introduction}

In the last decade, there was a tremendous progress in the research of the deep learning algorithms for virtually all domains and they are already pervasive through mobile phones, wearables, home robots.  
The progress in deep learning algorithms is the direct result of growing infrastructure in terms of available data and computing power. The Imagenet competition \cite{imagenet_cvpr09} was the reason that the deep learning algorithms became popular and dramatic improvements in algorithms proposed by different research communities. Impressive results in different domains \cite{SilverHuangEtAl16nature, adiwardana2020humanlike} were achieved using these modern techniques which were not expected to achieve at least for a decade. Motivated by this, the global giant companies are publishing more and more datasets in order to promote the algorithmic research. However, the preparation of such complex huge datasets, unavailability of computational infrastructure to all researchers as well as  delays for research to industrialization means a slow algorithmic improvement. Also, the datasets, even the high quality ones, are never true representative of the real data. 

Although this decade is expected to see more transition of this research into industrialization, it will also create new research questions related to privacy, interpretability/explanability, efficiency, ethics and adverserial attacks. Already the  European Union has passed the law ``The General Data Protection Regulation", to give all citizens right to protect their private data. This restricts the availability of the data not just to companies but also to researchers. This would also result into companies withholding themselves from publishing datasets as it might have some sensitive private data of users. This is harmful for the democratization of machine learning.

Federated learning provides a possibility to train machine learning models without the data leaving the edge devices. This eliminates all the logistical delays of algorithm development while providing the possibility to train the models on true data distribution. However, it suffers with latency problems because of encrypting the model parameters and sharing them with cloud. This operation is also very expensive in terms of computational power as the number of parameters are huge in most of the real world cases. Furthermore, the cloud distributes the aggregated parameters to each device which increases the chance of adversary attacks. Therefore, this paper proposes a new framework for federated learning that will simultaneously addresses the problem of privacy, efficiency and adversary attacks.

Section \ref{Background} describes the privacy preserving methodologies such as differential privacy, additive secret sharing and federated learning. The drawbacks of traditional federated learning are explained in Section \ref{SecDrawbacks} which are addressed by the FedCollabNN framework presented in Section \ref{FedCollabNN}. Section \ref{SecPATE} illustrates the private dataset generation method by edge models those are trained using FedCollabNN framework. Finally, \ref{SecResults} presents the implementation results of FedCollabNN framework on MNIST dataset and \ref{SecConclusion} summarize the conclusion and future tasks related to FedCollabNN framework.

Throughout this paper, the vectors are denoted by small bold letters while matrices are denoted by capital bold letters.
 
\section{Background} \label{Background}
There are many privacy techniques invented. However, this work makes use of combination of following three techniques to collaboratively train machine learning models without sharing their data, weights such that the resulting framework not only preserves privacy but it is also efficient and robust to adversarial attacks.

\tikzstyle{block} = [rectangle, draw, fill=blue!20, 
    text width=5em, text centered, rounded corners, minimum height=2em]
\tikzstyle{errorBlock} = [rectangle, draw, fill=red!20, 
     text centered, rounded corners, minimum height=2em]
\tikzstyle{arrow} = [ultra thick,<->,>=stealth]
\tikzstyle{place}=[circle,thick,draw=blue!75,fill=blue!20,minimum
                      size=6mm]
\tikzstyle{placeBlank}=[circle,thick,draw=white,fill=white,minimum
                      size=6mm]
\tikzstyle{OccGrid}=[rectangle,thick,draw=blue!75,fill=blue!20,minimum size=20mm]

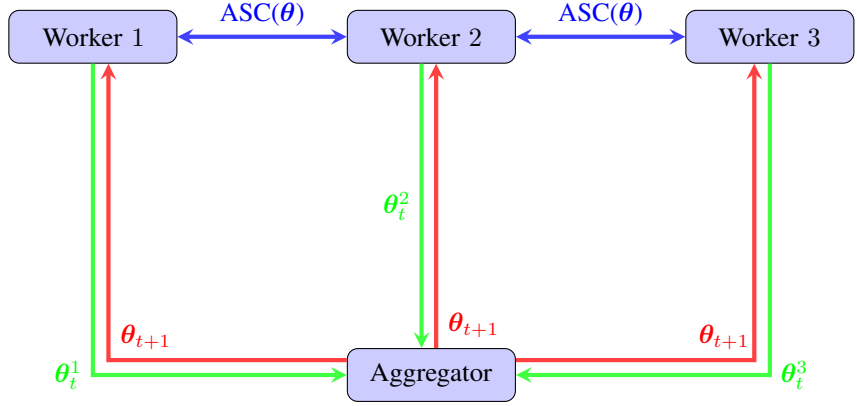
\begin{figure*}[!t]
\center
\begin{tikzpicture}[node distance = 2.5cm]

\node (W1) [block, text width=2 cm,align=center] {Worker $1$};
\node (W2) [block, right of = W1, xshift = 2 cm, text width=2 cm,align=center] {Worker $2$};
\node (Wn) [block, right of = W2,  xshift = 2 cm, text width=2 cm, align=center] {Worker $3$};

\node (Ag) [block, below of = W2,  yshift = -2 cm, text width=2 cm,align=center] {Aggregator};
\connectorHTextBluea[ultra thick,<->,>=stealth,draw=blue!75]{0.50}{W1}{W2};
\connectorHTextBluea[ultra thick,<->,>=stealth,draw=blue!75]{0.50}{W2}{Wn};
\connectorVTextGreena[ultra thick,->,>=stealth,draw=green!75]{0.50}{W2.250}{Ag.110};
\connectorHTextGreena[ultra thick,->,>=stealth,draw=green!75]{0}{W1.south}{Ag.west};
\connectorHTextGreenRighta[ultra thick,->,>=stealth,draw=green!75]{0}{Wn.270}{Ag};
\connectorVTextRedRighta[ultra thick,->,>=stealth,draw=red!75]{0}{Ag.10}{Wn.240};
\connectorVTextReda[ultra thick,->,>=stealth,draw=red!75]{0}{Ag.170}{W1.300};
\connectorVTextReda[ultra thick,->,>=stealth,draw=red!75]{0}{Ag.80}{W2.280};

\end{tikzpicture}
 \def\figurename{Figure}
\caption{Traditional Federated Learning Framework with 3 Workers. Green arrows indicate the transfer of model parameters from workers to aggregator while the red arrows represent transfer of aggregated parameters to workers. The blue arrow represent the additive secret sharing of parameters between workers.}
\label{FigTFL}
\end{figure*}

\subsection{Differential Privacy (DP)}
Differential privacy (DP) \cite{10.1561/0400000042} provides a methodology for providing strong privacy guarantees. It introduces statistical noise, that is significant enough to protect the privacy of an individual, but small enough that will not adversely impact the accuracy of the model. The formal definition is as follow:

A randomized algorithm $\mathcal{M}:\mathcal{D}\rightarrow R$ with domain $\mathcal{X}$ and range $R$, is $(\epsilon,\delta)$-differentially private if for any two adjacent training datasets $\mathcal{D}, \mathcal{D'} \subseteq \mathcal{X}$, which differ by at most one training point, and any subset of outputs $S \subseteq R$ it satisfies that:
\begin{equation}
Pr[\mathcal{M}(\mathcal{D}) \in S] \leq e^\epsilon Pr[\mathcal{M}(\mathcal{D'}) \in S] +\delta.
\end{equation}
The parameter $\epsilon$ is called privacy budget that defines an upper bound to privacy budget and $\delta$ is the probability with which this guarantee may not hold. Smaller budgets yield stronger privacy guarantees. 

Differential privacy comes in two different kinds which refer to the two different places that you can add noise. Local Differential Privacy adds noise to each individual data point. You can think of this as adding noise directly to the database or having individuals add noise to their own data before even putting it into the database. Global Differential Privacy adds noise to the output of the query on the database. This means that the database itself contains all the private information and that it’s only the interface to the data which adds the noise necessary to protect each individual’s privacy. Generally, local DP reduces accuracy of the model.

The application of differential privacy methods to the deep learning models by adding noise to the gradient in stochastic gradient descent algorithm \cite{Abadi2016DeepLW} opened a way to train differentially private generators as in \cite{DBLP:journals/corr/abs-1802-06739, Chen2018DifferentiallyPD, Tantipongpipat2019DifferentiallyPM, DBLP:journals/corr/abs-1904-02200}.

\subsection{Additive Secret Sharing (ADS)}
Secret sharing \cite{10.1145/359168.359176} refers to methods for distributing a secret amongst a group of participants, each of whom is allocated a share of the secret. The secret can be reconstructed only when a sufficient number, of possibly different types, of shares are combined together; individual shares are of no use on their own. Specifically, additive secret sharing allows multiple individuals to add numbers together without any person learning anyone else's inputs to the addition. 

\begin{algorithm}[tb]
   \caption{Traditional Federated Learning,  The $K$ clients indexed by $k$; $B^k$ is the set of minibatches for $k^{th}$ client, $\eta$ is the learning rate}
   \label{AlgTFL}
\begin{algorithmic}[1]
	\STATE {\bfseries Aggregator Executes:}
   \STATE {Initialize $\bm \theta_0$}
   \FOR {each round $t=1,2,\ldots$}
   \FOR {each client $k \in K$}
   \STATE {$\bm \theta^k_{t+1} \leftarrow$ WorkerUpdate$(k, \bm \theta^k_{t})$}
   \ENDFOR
   \STATE{$\bm \theta^k_{t+1}  \leftarrow$ FederatedAverage$(\bm \theta^k_{t+1} )$}
   \ENDFOR
   \item[]
   \STATE {\bfseries WorkerUpdate$(k, \bm \theta):$}
   \FOR {batch $b \in B^k$}
    \STATE  $\bm \theta \leftarrow \bm \theta - \eta\nabla\mathcal{L}(\bm \theta)$
   \STATE {AdditiveSharing$(\bm\theta)$} 
    \ENDFOR
    \STATE {return $\bm\theta$ to aggregator}
\end{algorithmic}
\end{algorithm}

\tikzstyle{block} = [rectangle, draw, fill=blue!20, 
    text width=5em, text centered, rounded corners, minimum height=2em]
\tikzstyle{errorBlock} = [rectangle, draw, fill=red!20, 
     text centered, rounded corners, minimum height=2em]
\tikzstyle{arrow} = [ultra thick,<->,>=stealth]
\tikzstyle{place}=[circle,thick,draw=blue!75,fill=blue!20,minimum
                      size=6mm]
\tikzstyle{placeBlank}=[circle,thick,draw=white,fill=white,minimum
                      size=6mm]
\tikzstyle{OccGrid}=[rectangle,thick,draw=blue!75,fill=blue!20,minimum size=20mm]

\begin{figure*}
\center
\begin{tikzpicture}[node distance = 2.5cm]

\node (W1) [block, text width=2 cm,align=center] {Worker $1$};
\node (W2) [block, right of = W1, xshift = 2 cm, text width=2 cm,align=center] {Worker $2$};
\node (Wn) [block, right of = W2,  xshift = 2 cm, text width=2 cm, align=center] {Worker $3$};

\node (Ag) [block, below of = W2,  yshift = -2 cm, text width=2 cm,align=center] {Aggregator};
\connectorHTextBlue[ultra thick,<->,>=stealth,draw=blue!75]{0.50}{W1}{W2};
\connectorHTextBlue[ultra thick,<->,>=stealth,draw=blue!75]{0.50}{W2}{Wn};
\connectorVTextGreen[ultra thick,->,>=stealth,draw=green!75]{0.50}{W2.250}{Ag.110};
\connectorHTextGreen[ultra thick,->,>=stealth,draw=green!75]{0}{W1.south}{Ag.west};
\connectorHTextGreenRight[ultra thick,->,>=stealth,draw=green!75]{0}{Wn.270}{Ag};
\connectorVTextRedRight[ultra thick,->,>=stealth,draw=red!75]{0}{Ag.10}{Wn.240};
\connectorVTextRed[ultra thick,->,>=stealth,draw=red!75]{0}{Ag.170}{W1.300};
\connectorVTextRed[ultra thick,->,>=stealth,draw=red!75]{0}{Ag.80}{W2.280};

\end{tikzpicture}
 \def\figurename{Figure}
\caption{FedCollabNN Framework with 3 Workers. Green arrows indicate the transfer of loss from workers to aggregator, while the red arrows represent transfer of aggregated loss to workers. The blue arrow represent the additive secret sharing of loss values between workers.}
\label{FigFedCollabNN}
\end{figure*}
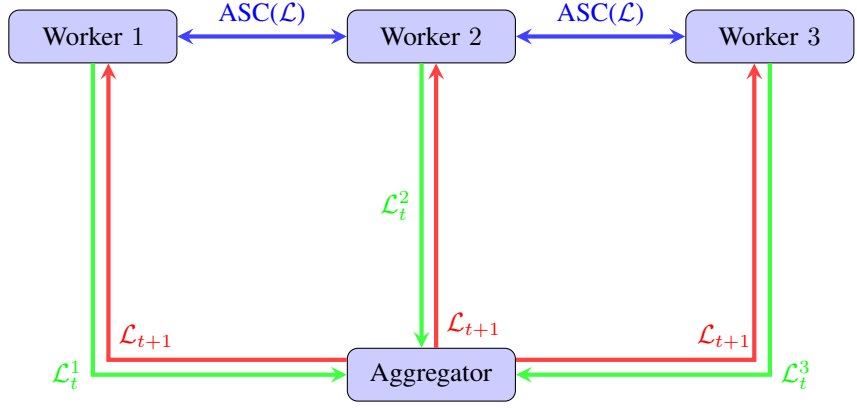

\subsection{Federated Learning (FL)}
Standard machine learning approaches require centralizing the training data on one machine or in a datacenter. Federated Learning (FL) \cite{45648} enables collaborative learning a shared prediction model while keeping all the training data on device.  The algorithm for traditional FL is described in Alg.\,\ref{AlgTFL} and graphically described in Fig.\,\ref{FigTFL}. A worker device downloads the current model with parameters $\bm\theta_t$  in $t^{th}$ iteration (line 5) from cloud, improves it by learning from data on the device (line 11), and then summarizes the changes as a small update. Only this update to the model is sent to the cloud, using encrypted communication such as additive secret sharing (line 12), where it is immediately averaged with other user updates to improve the shared model (line 7). All the training data remains on your device, and no individual updates are stored in the cloud.

The training goal for traditional federated learning is minimizing the cost function $J$ such that
\begin{equation}
J = \argmin_{\bm\theta^1} f(\bm x^1) +  \ldots + \argmin_{\bm\theta^K} f(\bm x^K).
\end{equation} 
The parameters for models on each worker become same after each iteration. Therefore, the cost function becomes
\begin{equation}
J = \argmin_{\bm\theta} f(\bm x^k),
\end{equation} 
where $k=1,2,\ldots,K$ and $\bm\theta = \bm\theta_1 =\bm\theta = \ldots = \bm\theta_K$.

\section{Drawbacks of Traditional FL Framework} \label{SecDrawbacks}
Although traditional FL provides a nice framework for providing the privacy to users and is considerably efficient than traditional centralised machine learning framework, it suffers with few drawbacks. 

First and most important issue is of latency. Even the simplest deep learning algorithms usually consist of thousands of parameters. Sending these parameters to aggregator and  receiving updates is very expensive task. Additionally, the model parameters have to be encrypted and decrypted. The edge devices do not have similar computational power and resources. Therefore, the usage of large models is difficult leading to low performance. 

Secondly, as the model parameters for all devices are same, an adversary can take part in the training process and influence the results. He also gets an access to the model parameters using which he can devise adverserial attacks to find user memberships.

\section{FedCollabNN Framework} \label{FedCollabNN}
FedCollabNN is a collaborative learning framework that achieves a low latency training by changing the goal from local optimization of machine learning models at edge to the global optimization of  machine learning models on all edge devices. Instead of locally backpropagating the loss on each device and aggregating parameters over the cloud, FedCollabNN aggregates the loss from all devices, averages them and backpropagate it to all models. This requires only sending and receiving a single scalar value to and from the cloud to each device. This reduces the latency dramatically and is unaffected by the size of model. Also, it does not force all devices to have same models or parameters which make them robust to adverserial attacks as adversaries do not have an access to the model parameters, data and labels. The optimization goal for FedCollabNN is described as  
\begin{equation}
J = \argmin_{\bm\phi} \mathcal{L}^{avg},
\end{equation}
where $\bm\phi = \{\bm\theta^1, \bm\theta^2, \ldots, \bm\theta^K\}$ and $\mathcal{L}^{avg}$ is the average loss calculated by the aggregator. The algorithm is similar to traditional federated learning with only difference is that the additive secret sharing and aggregation is done with loss values instead of model parameters. The algorithm is describedin Alg.\,\ref{AlgFedCollabNN} and the Fig.\,\ref{FigFedCollabNN} shows the FedCollabNN framework with 3 workers.

\begin{algorithm}[tb]
   \caption{FedCollabNN Framework.  The $K$ klients indexed by $k$; $B$ is the local minibatch size, $\eta$ is the learning rate}
   \label{AlgFedCollabNN}
\begin{algorithmic}[1]
	\FOR {$k=1,2,\ldots,K$}
	\STATE {Initialize $\bm\phi = \{\bm\theta^1, \bm\theta^2, \ldots, \bm\theta^K\}$}
	\ENDFOR
	\STATE {\bfseries Aggregator Executes:}
   \FOR {each round $t=1,2,\ldots$}
   \FOR {each client $k \in K$}
   \STATE {${\mathcal{L}}^k_{t+1} \leftarrow$ WorkerLossCalculate$(k, {\mathcal{L}}_{t} )$}
   \ENDFOR\\
   ${\mathcal{L}}_{t+1}  \leftarrow \sum_{k=1}^K \mathcal{L}^k_{t+1}$
   \ENDFOR
   \item[]
   \STATE {\bfseries WorkerLossCalculate$(k,  {\mathcal{L}} )$}
    \STATE  $\bm{\theta}^{k} \leftarrow \bm{\theta}^{k} - \eta\nabla\mathcal{L}(\bm{\theta}^{k})$
   \STATE {$\mathcal{L} \leftarrow$ ForwardComputation($\bm\theta^{k}$)}
    \STATE {AdditiveSharing$(\mathcal{L})$} 
    \STATE {return $\mathcal{L}$ to aggregator}
\end{algorithmic}
\end{algorithm}

\section{Private Data Generation using FedCollabNN Framework} \label{SecPATE}
The PATE framework methodology \cite{Papernot2018ScalablePL} is used to generate the data that protects the privacy. The core of PATE framework is based on differential privacy concept that if the dataset contains a particular amount of private information, no post-processing can divulge more information than that in the dataset. PATE transfers knowledge from an ensemble of teacher models trained on partitions of the private data to a student model. 

Each teacher is a model trained independently on a subset of the data whose privacy one wishes to protect. Training each teacher on a partition of the sensitive data produces different models solving the same task. At inference, teachers independently predict labels.
 To provide rigorous guarantees of differential privacy, the aggregation mechanism of the original PATE framework counts votes assigned to each class, adds carefully calibrated Laplacian noise to the resulting vote histogram, and outputs
the class with the most noisy votes as the ensemble’s prediction. PATE’s final step involves the training of a student model by knowledge transfer
from the teacher ensemble using access to public—but unlabeled—data. As the aim of this work is to just generate privacy preserving data, the models learned on each worker are taken as teacher models. A public dataset with just inputs are sent to each worker and the PATE framework procedure can be followed to generate corresponding labels which can be used for training student models, statistical analysis, etc.

\section{Results} \label{SecResults}
The implementation of FedCollabNN framework for MNIST using PySyft library \cite{1111} can be found \href{https://github.com/Amit507017/FedCollabNN}{here}.  The simulations are done with 32000 training and 10000 test data. The model architectures and other training parameters can be see on the Github repository. A number of experiments are necessaryy to understand the advantages and disadvantages for this framework in comparison to other approaches. Few results are presented below.

\subsection{Changing the Number of Collaborators} 
\begin{figure}
\includegraphics[scale=0.6]{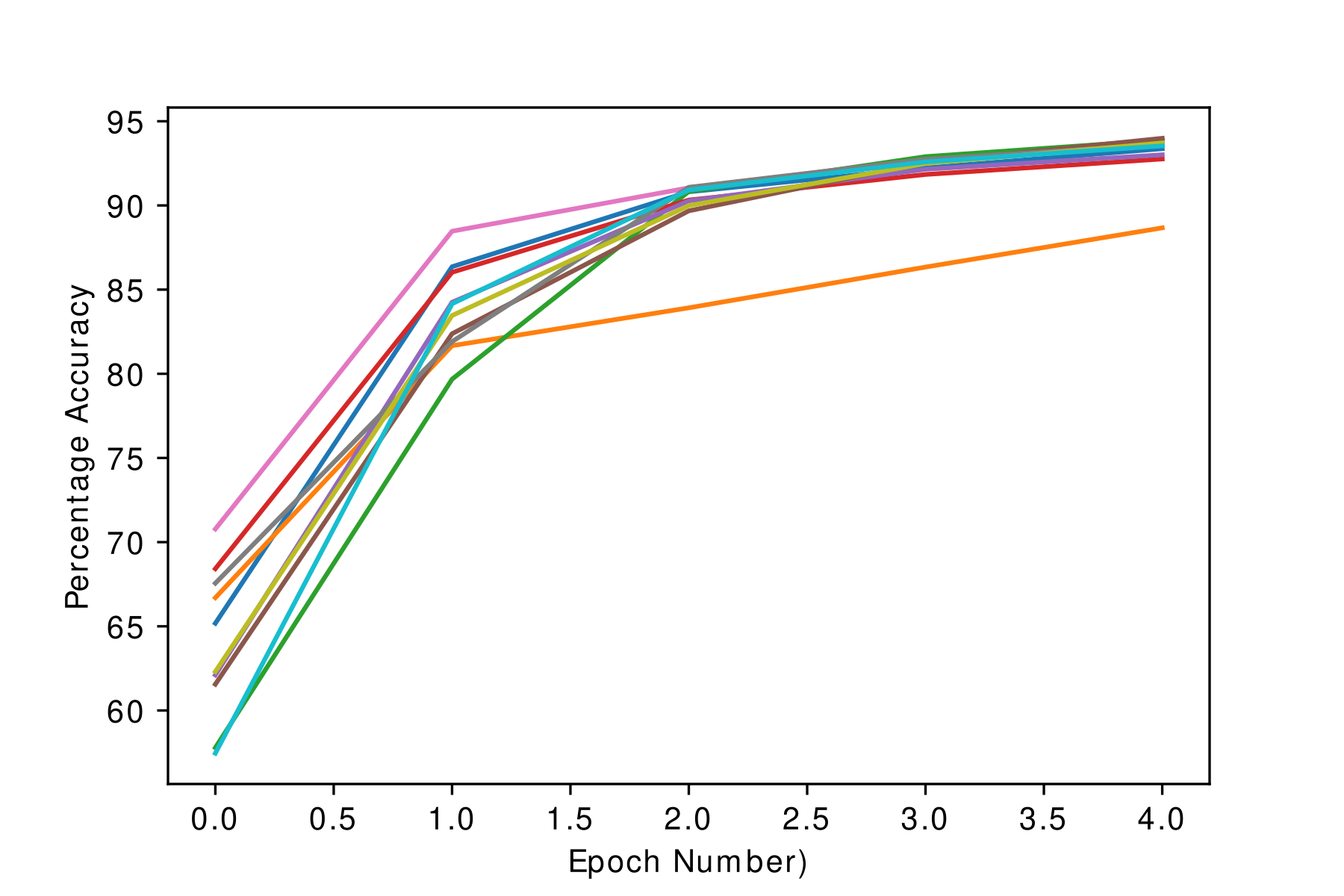} 
\vspace{-0.6 cm}
\caption{ Accuracies using FedCollabNN Framework with 10 Workers}
\label{Fig10Workers}
\end{figure}

\begin{figure}
\includegraphics[scale=0.6]{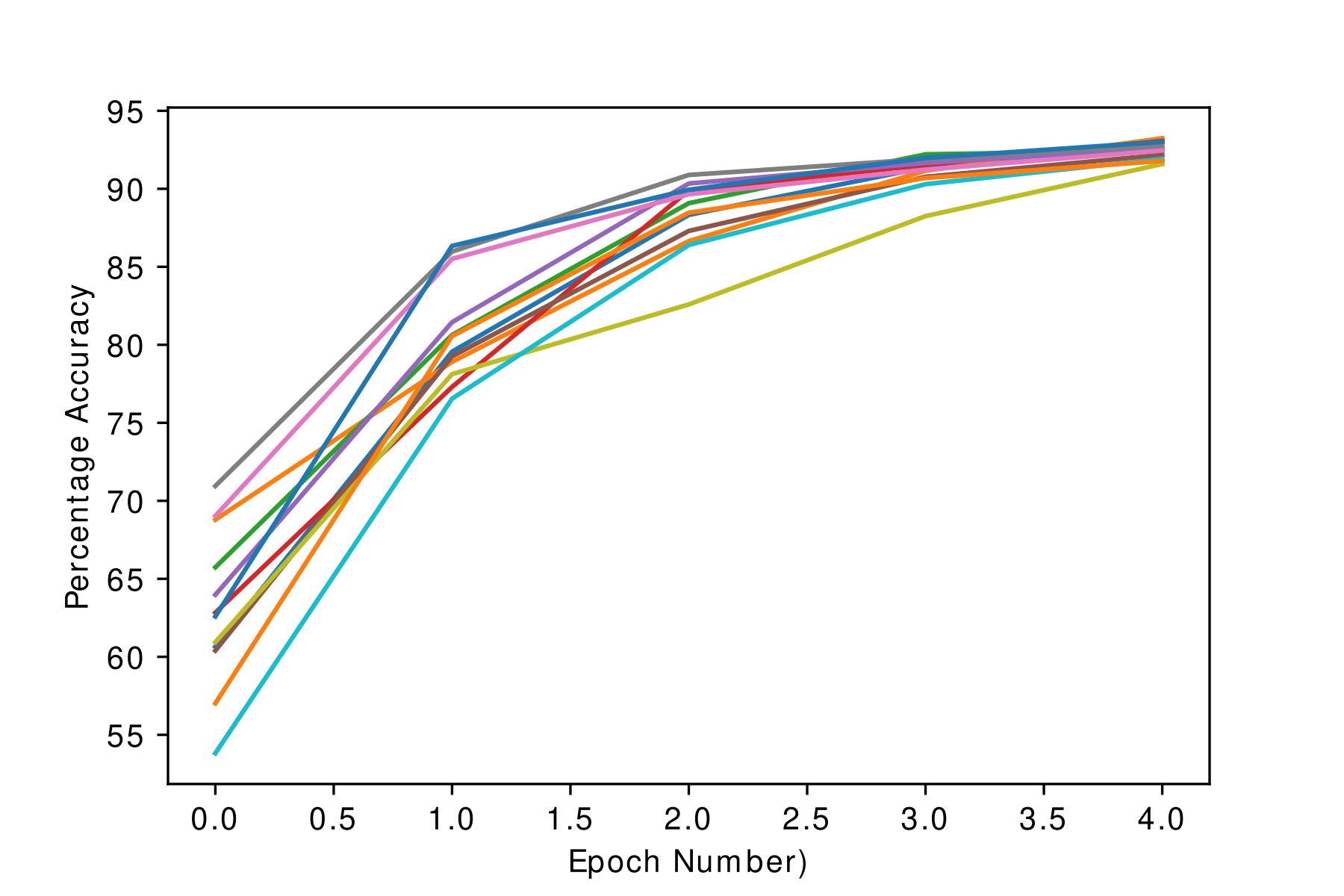} 
\vspace{-0.6 cm}
\caption{ Accuracies using FedCollabNN Framework with 12 Workers}
\label{Fig12Workers}
\end{figure}

\begin{figure}
\includegraphics[scale=0.6]{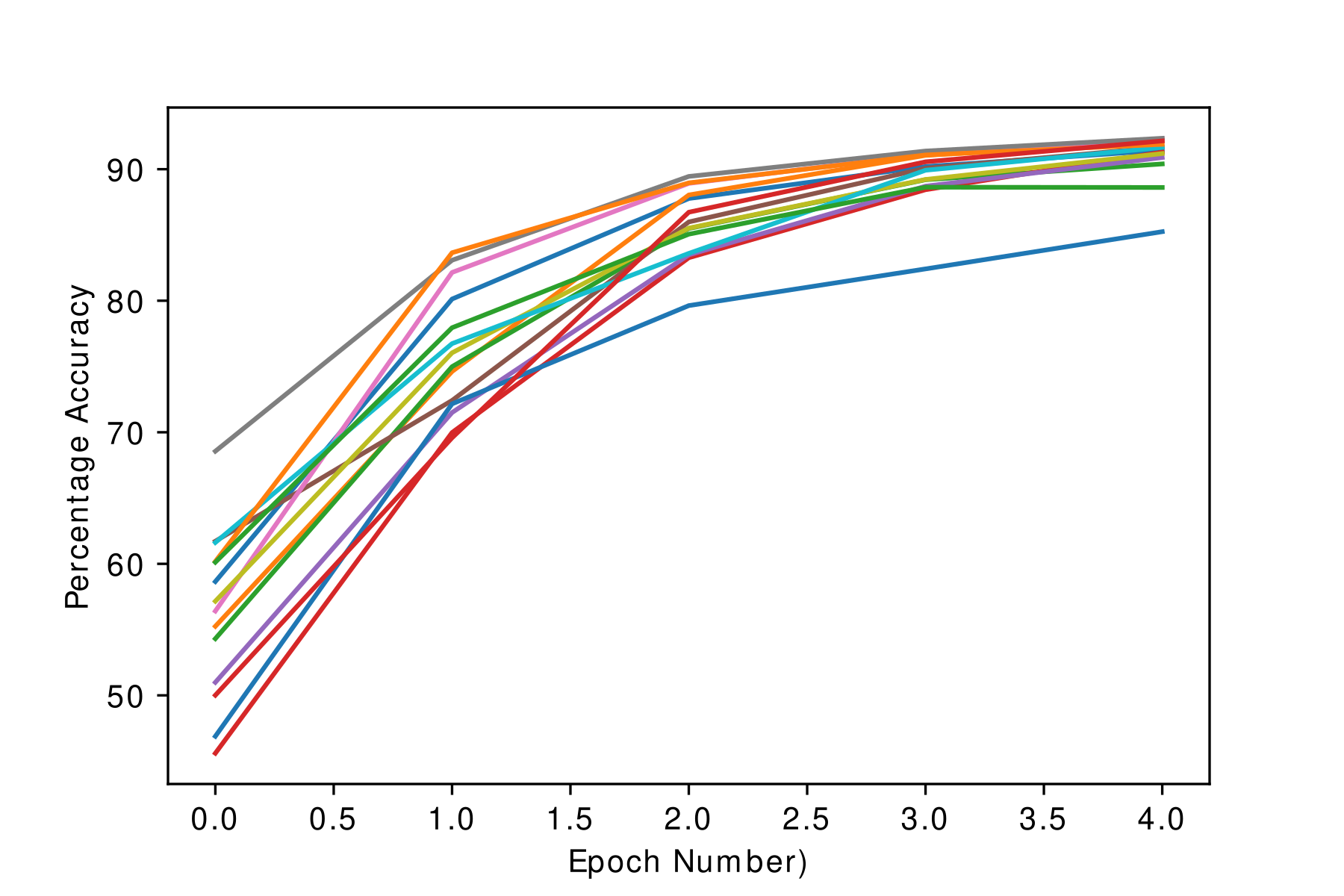} 
\vspace{-0.6 cm}
\caption{ Accuracies using FedCollabNN Framework with 14 Workers}
\label{Fig14Workers}
\end{figure}

\begin{figure}
\includegraphics[scale=0.6]{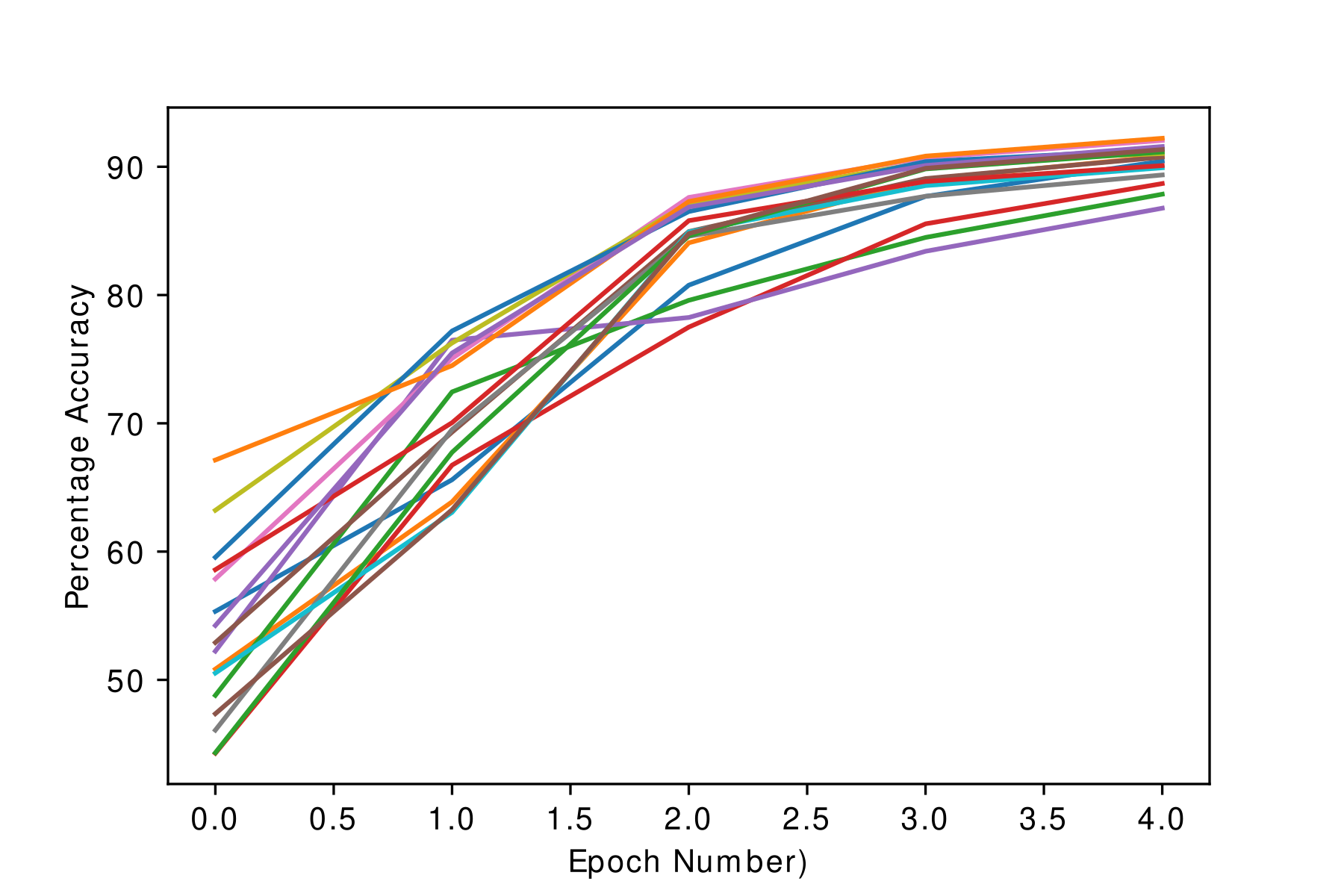} 
\vspace{-0.6 cm}
\caption{ Accuracies using FedCollabNN Framework with 16 Workers}
\label{Fig16Workers}
\end{figure}

\begin{figure}
\includegraphics[scale=0.15]{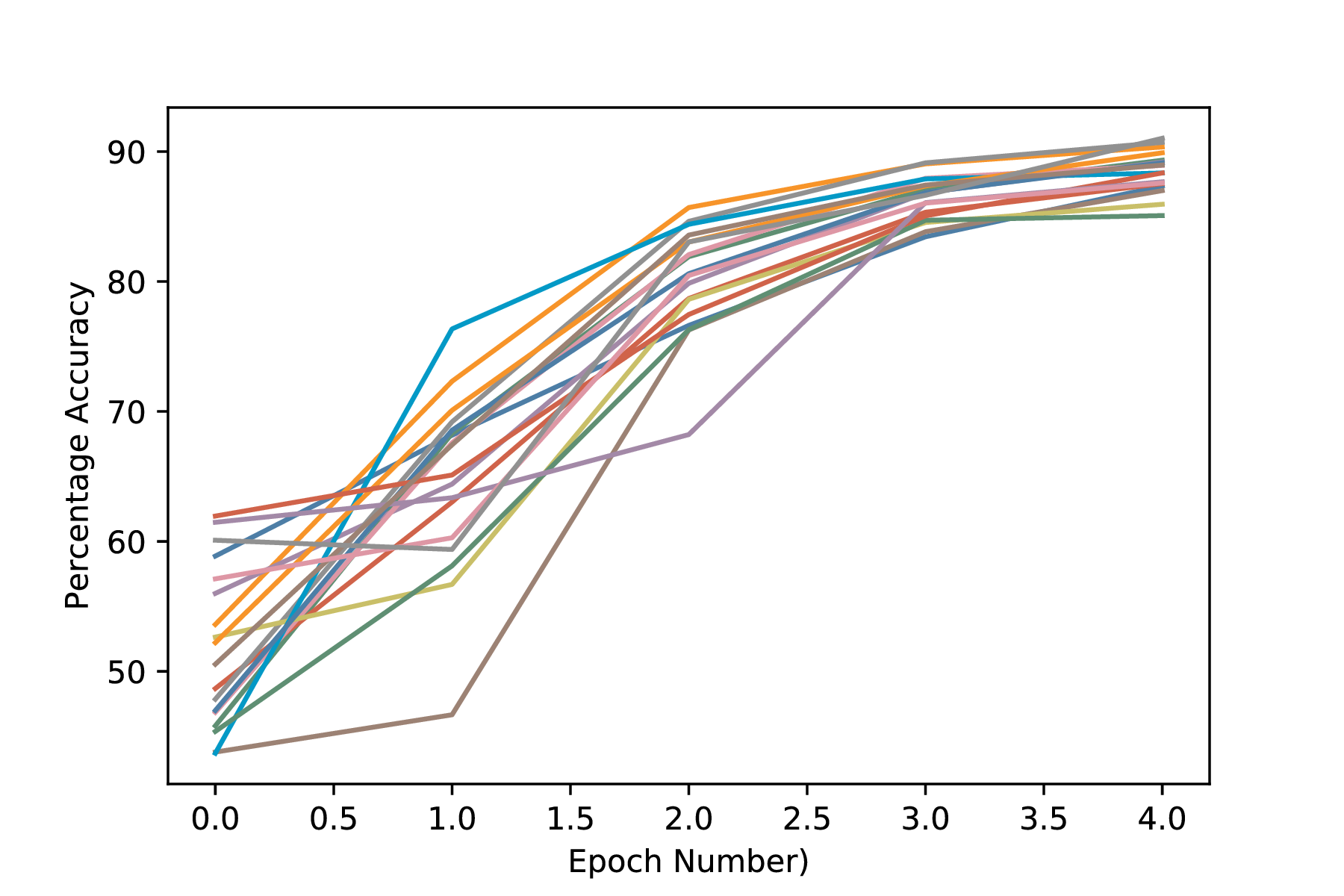} 
\vspace{-0.6 cm}
\caption{ Accuracies using FedCollabNN Framework with 18 Workers}
\label{Fig18Workers}
\end{figure}

\begin{figure}
\includegraphics[scale=0.15]{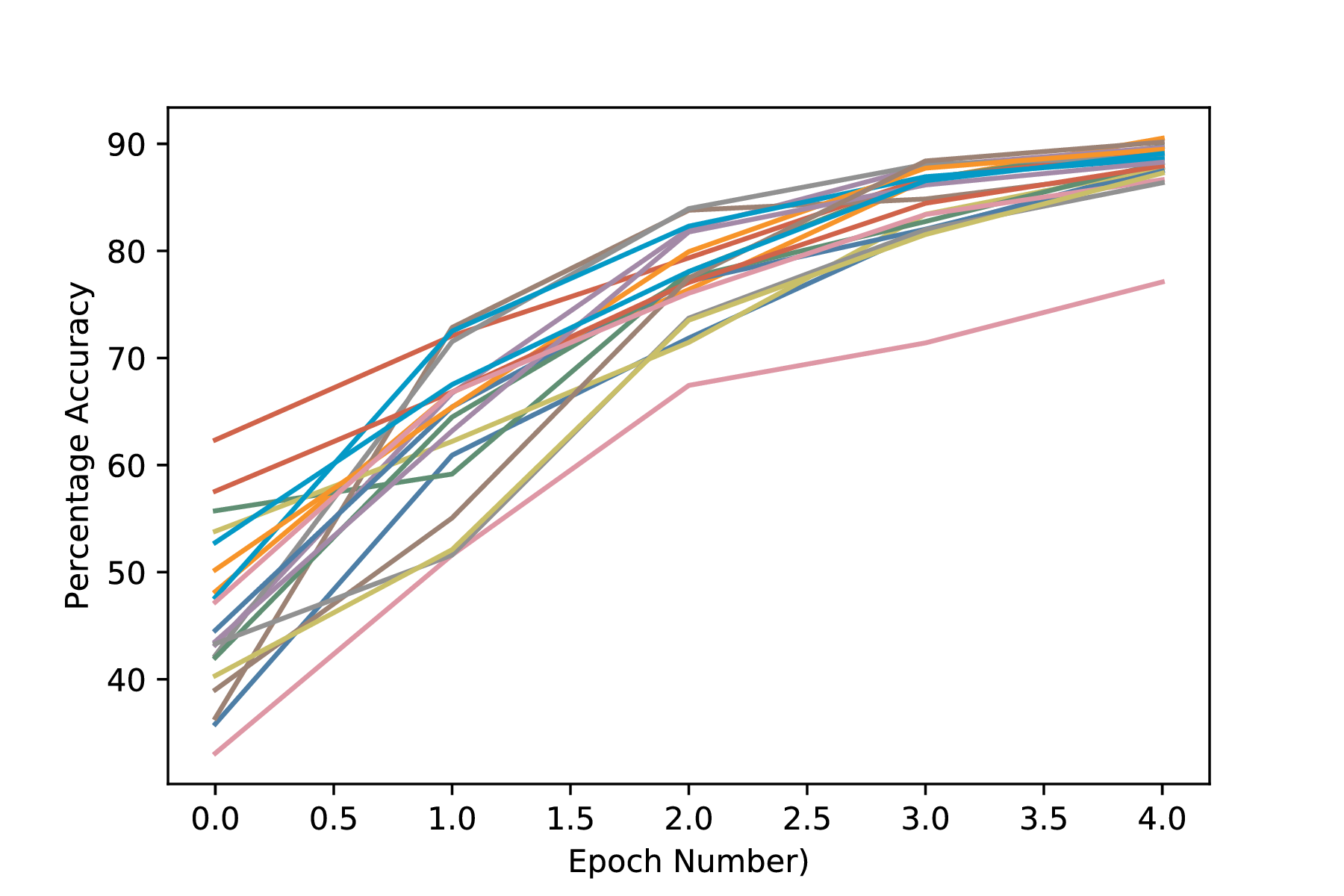} 
\vspace{-0.6 cm}
\caption{ Accuracies using FedCollabNN Framework with 20 Workers}
\label{Fig20Workers}
\end{figure}

As the parameters for all models for workers are different, it increases the parameter space dramatically and it could hamper training performance as the number of workers increases. Fig.\,\ref{Fig10Workers}, \ref{Fig12Workers}, \ref{Fig14Workers}, \ref{Fig16Workers}, \ref{Fig18Workers}, \ref{Fig20Workers} refers to the test accuracies for each model of the worker when they learn collaborativelyy with FedCollabNN framework over the five Epochs. General trend in all scenarios show that the workers learn collaboratively and each worker improve model accuracy with epochs. The accuracies decrease slightly with the increase in the number of workers. However, the tail of each curve still shows an upward trend which indicates better accuracies can be achieved with further training and perhaps more epochs are necessary with large number of workers. 


\subsection{Inference using Aggregator}
\begin{figure}
\includegraphics[scale=0.6]{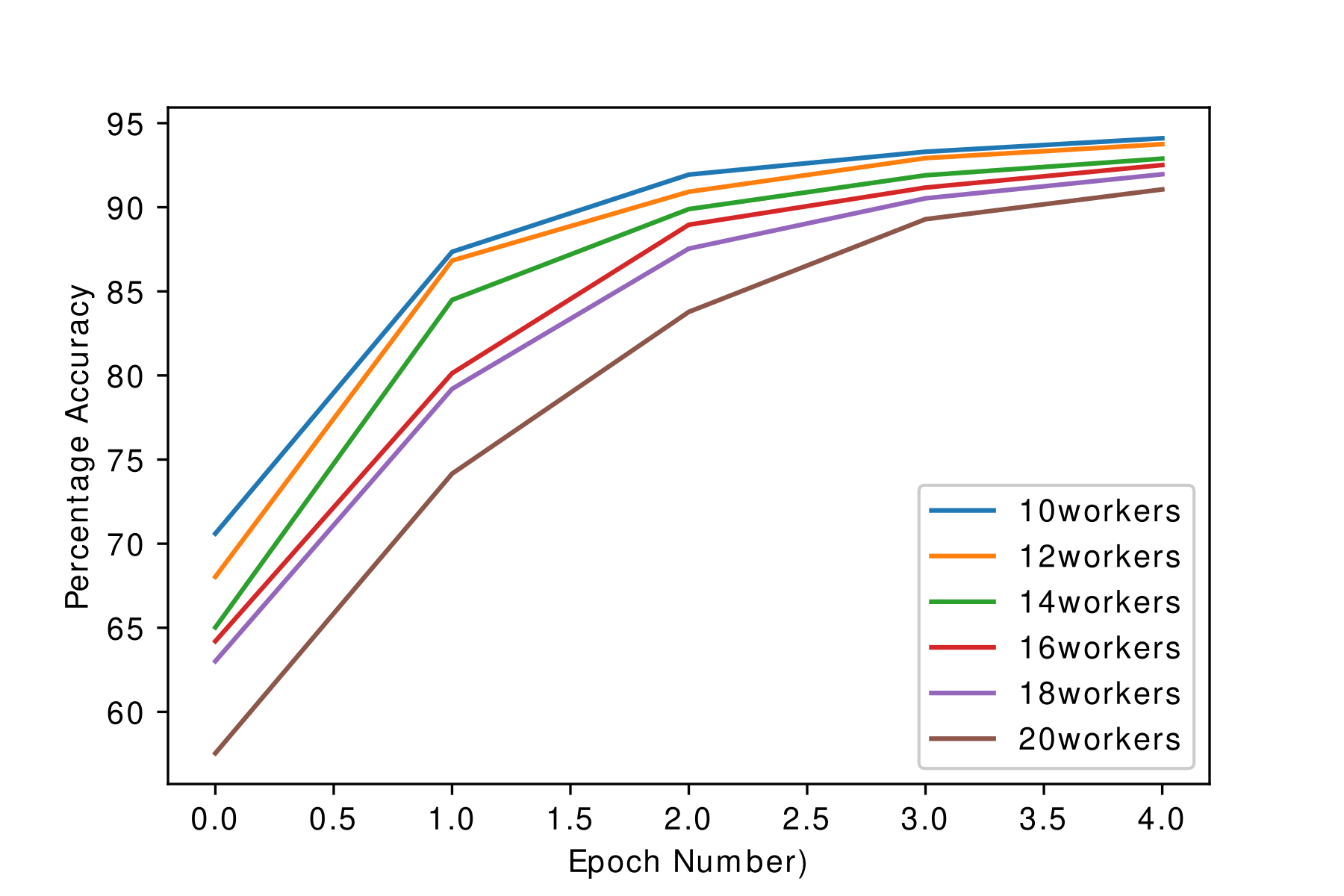} 
\vspace{-0.6 cm}
\caption{Aggregated Test Accuracies with different Number of Workers}
\label{FigTestAggregate}
\end{figure}

The test accuracy for aggregated result is shown in Fig.\,\ref{FigTestAggregate}. It shows as the number of workers increases, the aggragated test accuracy decreases slightly. However, the trend still shown that with more epochs the accuracy will converge to similar values. Another thing to notice here is that the aggragated test accuracy is always at least little higher than the individual worker's test accuracy. This means that the generated data not only preserves the privacy but also is of high quality. 

\section{Conclusion and Outlook} \label{SecConclusion}
This paper proposed a FedCollabNN framework, a variant of federated learning algorithm, that does privacy preserving collaborative learning by training models on edge without transferring the data from edge. The latency is much low compared to traditional federated learning as it shares only loss, i.\,r., a single scalar value, with aggregator. Also, each worker have model with different parameters that makes it less prone to adversarial attacks. All these models in combination with a global differential privacy mechanism provides a methodology for data generation for further analysis. The simulation results on MNIST dataset show promising results.

Many scenarios still need to be simulated and many improvements need to made in the framework such that the accuracy of all worker models and aggregated model increases. The framework considers fixed number of workers from start to end. The methodology for adding and subtracting workers to the training process and their influence on model accuracies need to be analysed.

As the parameters of the worker models are different, it also provides a possibility of training personalised models for workers. In order to decrease the parameter space, partly parameters can also be shared. 

\bibliographystyle{plainnat}
\bibliography{Ref}

\end{document}